\documentclass{svproc}
\usepackage{url}

\usepackage{graphicx} 

\begin{document}
\mainmatter              
\title{Prototype-based classifiers in the
presence of concept drift: A modelling framework}
\titlerunning{Prototytpe-based classifiers under drift}  
%
\author{
Michael Biehl\inst{1} 
\and Fthi Abadi\inst{1} 
\and Christina G{\"o}pfert \inst{2}
\and Barbara Hammer \inst{2}}
\authorrunning{Michael Biehl et al.} 
%
\tocauthor{Michael Biehl, Fthi Abadi, Christina G{\"o}pfert and Barbara Hammer} 
\institute{University of Groningen,
Bernoulli Institute for Mathematics, Computer Science and Artificial Intelligence, P.O. Box 407, NL-9700 AK Groningen, The Netherlands\\
\email{m.biehl@rug.nl, fthialem@gmail.com}\\
WWW home page:
\texttt{http://www.cs.rug.nl/\~{}biehl}
\and
Bielefeld University,
Center of Excellence - Cognitive Interaction Technology, CITEC,
Inspiration 1, D-33619 Bielefeld, Germany 
\\
\email{bhammer\{cgoepfert\}@techfak.uni-bielefeld.de}}
\maketitle              

\begin{abstract}
We present a modelling framework for the 
investigation of 
proto\-type-based classifiers in 
non-stationary environments.
Specifically, we study
Learning Vector Quantization (LVQ) systems
trained from a stream of high-dimensional, 
clustered data. 
We  consider standard 
winner-takes-all updates known as LVQ1. 
Statistical properties of the input data
change on the time scale 
defined by the training process.  We apply 
analytical methods borrowed from statistical physics 
which have been used earlier 
for the exact description of  learning in 
stationary environments. 
The suggested framework facilitates the 
computation of learning curves in the 
presence of virtual and real concept drift. 
Here we focus on time-dependent class bias 
in the training data. 
First results demonstrate that, while 
basic LVQ algorithms 
are suitable for the training in
non-stationary environments, 
\textit{weight decay} as an explicit 
mechanism of forgetting does not improve the
performance under the considered drift processes. 
\keywords{LVQ, concept drift,
weight decay, supervised learning}
\end{abstract}

\section{Introduction}

\label{intro} 

The topic of learning under \textit{concept drift} is 
currently attracting increasing interest in the 
machine learning community. Terms like 
\textit{lifelong learning}
or \textit{continual learning} have been coined in 
this context \cite{driftreview}.

In the standard set-up, machine learning 
processes \cite{Hastie} are conveniently
separated into two stages: 
In the so-called {\it  training phase,} a 
hypothesis or model of the data
is inferred from a given set of example data. 
Thereafter, this hypothesis can be applied
to novel data in the {\it working phase,} 
e.g. for the purpose  of classification or 
regression. 
Implicitly, the training data is assumed to 
represent the target task faithfully also 
after completing the training phase:
Statistical properties of
the data and the task itself should not 
change in the working phase.

Frequently, however, 
the separation of training and working 
phase appears artificial or unrealistic, for 
instance in  human or other biological learning  
processes \cite{Cetraro2}.  
Similarly,  in technical contexts, 
training data is often available in the form 
of non-stationary data streams, e.g. 
\cite{driftreview,Losing,Ditzler,Joshi,Ade}. 
Two major types of non-stationary
environments have been discussed in the 
literature: 
In {\it virtual drifts,} statistical properties 
of the available training data are 
time-dependent,  while the actual 
target task remains unchanged.  
Scenarios in which the target itself, e.g.\ the 
classification or regression scheme, 
changes in time are referred to
as {real drift} processes. 
Frequently both effects coincide, further
complicating the detection and handling of the 
drift. 
  
 In general, the presence of drift
 requires the \textit{forgetting} of 
 older 
 information while  the system is adapted to 
 recent example data. 
 The design of efficient forgetful
 training schemes
 demands a thorough theoretical understanding
 of the relevant phenomena.
To this end, the development
of suitable modelling frameworks is 
instrumental.
 Overviews of earlier work and recent 
 developments
 in the context of non-stationary learning
 environments can be found   
 in e.g.\ 
 \cite{driftreview,Losing,Ditzler,Joshi,Ade}. 
 
Here, we study a basic model of
learning in a non-stationary environment. 
In the proposed 
framework we can address both
virtual and real drift processes. 
An example study of the latter has 
been presented in \cite{preprint}, recently,
where the specific case 
of random displacements of cluster centers in a
bi-modal input distribution was considered. 
Here,  however, the focus is on the study of
localized, but 
explicitly time-dependent densities of  
high-di\-men\-sio\-nal 
inputs in a stream of training examples. 
More specifically, we consider Learning Vector 
Quantization (LVQ) as a   
 prototype-based framework for classification
 \cite{kohonen1,nova,wires}.
 LVQ systems are most frequently trained in an online 
 setting by presenting a sequence of single examples
 for iterative adaptation \cite{nova,wires,jmlr}. 
 Hence, LVQ should constitute a natural tool 
 for incremental learning
 in non-stationary environments 
  \cite{Losing}.
 
    Methods developed
    in statistical physics facilitate
    the mathematical
    description of the 
    training dynamics in terms of typical 
    learning curves. 
    The statistical mechanics of on-line 
    learning has helped to gain insights into 
 the typical behavior of various learning systems, see e.g.  
 \cite{cambridge,engel,revmod} and references therein. 
 
Clustered densities of data, similar to the one 
considered here, 
have been studied in the modelling of
unsupervised learning and
supervised perceptron training, see e.g.\ 
\cite{freking,sompo,marangi}.
In particular, online LVQ in stationary 
situations was analysed 
in \cite{jmlr}. 
Simple models of concept drift have been
studied before
within the statistical physics theory of the perceptron: 
Time-varying linearly separable classification rules
were considered in \cite{drifting2,caticha2}.
  
  We focus on the question 
  whether LVQ learning schemes are able to cope 
  with drift in characteristic model situations 
  and whether extensions like weight decay
  can further improve 
  the performance of LVQ in such settings.

\section{Models and Methods} 
First, we introduce Learning Vector 
Quantization 
for classification tasks with emphasis on the
basic LVQ1 scheme.  We propose a 
model density of data,  which was previously
investigated in the mathematical 
analysis of LVQ training in 
stationary environments. 
Finally, we extend the approach to 
the presence of 
concept drift and consider
{\it weight decay} as an explicit mechanism of
forgetting.

\subsection{Learning Vector Quantization} \label{LVQ} 
The family of LVQ 
algorithms is widely 
used for practical classification problems
\cite{nova,wires}. 
The popularity of LVQ is due to a number 
of attractive features:
It is quite easy to implement, very
flexible and intuitive. 
Multi-class problems can
be handled in a  natural way by introducing
at least one prototype per class. 
The actual classification scheme is 
most frequently based on Euclidean metrics
or other simple measures, which 
quantify the distance
of data (inputs, feature vectors)
from the class-specific prototypes obtained from
the training data. Moreover, 
in contrast to many other methods,
LVQ facilitates direct 
interpretation since the
prototypes are defined in the 
same space as the data \cite{nova,wires}.

\subsubsection{Nearest Prototype Classifier}\ \\
We restrict the analysis to 
the simple case of 
only one prototype per class in binary classification problems. Hence we consider
two prototypes $\vec{w}_S  \in I\!\!R^N$ 
with the subscript $S=\pm1$ (or $\pm$ for short) 
indicating the represented class of data. 
The system parameterizes  
a Nearest Prototype Classification (NPC) scheme
in terms of a distance measure
$d(\vec{w},\vec{\xi}):$ Any given  input 
$\vec{\xi} \in I\!\!R^N$ is 
assigned to the class label $S=\pm1$ of the closest prototype.
A variety of distance measures have  been
used in  LVQ, enhancing the flexibility of the approach even further \cite{wires,nova}. 
Here, we restrict 
the analysis to the  - arguably - simplest 
choice: the 
(squared) Euclidean measure
$ 
d(\vec{w}, \vec{\xi})= (\vec{w} - \vec{\xi})^2.
$

\subsubsection{The LVQ1 algorithm} \ \\
A sequence of single example data $\{ \vec{\xi}^{\, \mu}, \sigma^\mu \}$ is 
presented to the LVQ system  in the on-line
training process
\cite{kohonen1,jmlr}: 
At a given 
time step $\mu \, = \, 1,2,\ldots, $ the
feature vector $\vec{\xi}^{\, \mu}$  is presented
 together with the class label
$\sigma^\mu=\pm1$. 

Incremental 
LVQ updates are of the quite general form (see \cite{jmlr})
\begin{equation}  \label{generic}
  \vec{w}_S^\mu = \vec{w}_S^{\mu-1} + 
  \Delta \vec{w}_S^\mu 
 \mbox{~with~}
 \Delta \vec{w}_S^\mu  = 
  \frac{\eta}{N} f_S\left[d_{+,-}^{\mu},\sigma^\mu,\ldots\right]
 \left(\vec{\xi}^\mu - \vec{w}_S^{\mu-1}\right),
\end{equation}
where the vector $\vec{w}_S^\mu$ denotes the prototype after 
presentation
of $\mu$ examples 
and the constant learning rate $\eta$ is 
scaled with
the input dimension $N$. 
The actual algorithm is defined through the
so-called {\it modulation function}  
$f_S[\ldots]$, 
which  typically depends on the labels of the 
data and prototypes and on
the relevant distances of the input
from the prototype vectors.

Taking over the NPC concept, the LVQ1 training
algorithm \cite{kohonen1}
modifies only the  the so-called
{\it winner,} i.e. the prototype closest to
the current training input.  
The LVQ1 update for two competing 
prototypes corresponds to Eq.\ (\ref{generic}) with
\begin{equation} \label{LVQ1f}
f_S[d_+^\mu,d_-^\mu,\sigma^\mu] = 
 \Theta\left(d_{-S}^\mu - d_{+S}^\mu\right) 
 S  \sigma^\mu 
\mbox{~~where~}
\Theta(x) = 1 \mbox{~if~} x>0
 \mbox{~and~} 0 \mbox{~else.}
\end{equation}
The prefactor 
$S \, \sigma^\mu = \pm 1$ specifies 
the direction of the update: the 
\textit{winner} is moved 
towards the 
presented feature vector
if it carries the same class label, while its distance from the data point
is further increased if the labels disagree. 

\subsection{The dynamics of LVQ} \label{dynamics}
Statistical physics based methods 
have been used very successfully 
in the analysis of various learning systems 
\cite{engel,revmod}. 
The methodology is complementary to 
other frameworks of computational learning theory
and aims at the description of
typical learning dynamics
in simplifying model scenarios.
Frequently, the approach is based on
the assumption 
that a sequence of statistically independent, randomly generated $N$-dimensional 
input vectors is presented to the learning 
system. 
Further 
simplifications and the consideration of the 
thermodynamic limit $N\to\infty$ 
facilitate the mathematical
representation of the learning dynamics
by ordinary differential equations (ODE)
and the computation of {\it learning curves}.

Here, we extend earlier investigations of 
LVQ training in the
framework of a simplifying model situation \cite{jmlr}: 
High-di\-men\-sio\-nal training samples are 
generated independently according to a 
mixture of two overlapping Gaussian clusters. 
The input vectors are labelled according to 
their cluster membership and presented 
to the LVQ1 system, sequentially.  
Similar models have been investigated
in the context of other learning scenarios, 
 see for instance  
 \cite{freking,sompo,marangi}.

\subsubsection{The Data} \ \\
We consider random input vectors $\vec{\xi} \in I\!\!R^N$ which are
generated independently according to a
bi-modal distribution of the form \cite{jmlr}
\begin{equation} \label{prob}
 P(\vec{\xi}) = \sum_{\sigma=\pm1}\! p_\sigma \,  P(\vec{\xi}\!\mid\!\!\sigma) \mbox{~with~}
  P(\vec{\xi}\!\mid\!\sigma)  = \frac{1}{(2 \pi v_\sigma)^{\frac{N}{2}}}  \, 
 \exp \left[-\frac{1}{2 \, v_\sigma} \left( \vec{\xi} - \lambda 
\vec{B}_\sigma \right)^2 \right].
\end{equation}
The class-conditional densities $P(\vec{\xi}\mid\sigma\!=\!\pm1)$ represent
isotropic, spherical clusters with 
variances $\, v_\sigma$ and
means  given by $\lambda \, \vec{B}_\sigma$.
Prior weights of these Gaussian clusters 
are denoted by
$p_\sigma$ with $p_+ + p_- =1$.
For simplicity, we assume that the 
vectors $\vec{B}_\sigma$  are normalized,   
 $\vec{B}_+^{\, 2}=\vec{B}_-^{\, 2}=1$,
 and orthogonal with 
 $\vec{B}_+ \cdot \vec{B}_- =0$.
The target classification for each input
is given by its
class-membership $\sigma=\pm1$. The 
problem is not linearly separable since
the clusters overlap. 

Conditional averages over  
$P(\vec{\xi}\!\mid\!\sigma)$  will be denoted as 
$\left\langle \cdots \right\rangle_\sigma $, 
while
mean values of the form  $\langle \cdots 
\rangle = 
\sum_{\sigma=\pm1} \, p_\sigma \, \left\langle\cdots\right\rangle_\sigma $
are defined for the full density (\ref{prob}). 
In a particular cluster $\sigma$, input
components  $\xi_j$  are 
statistically independent and display 
the variance $v_\sigma$.
We will use, e.g., 
 the following (conditional) averages:
\begin{equation}  \label{xinorm}
 \left\langle \xi_j \right\rangle_\sigma = 
 \lambda (\vec{B}_\sigma)_j, \mbox{~~}
 \langle \vec{\xi}^{2} \rangle_\sigma  
   =  v_\sigma N  + \lambda^2,
 \mbox{~~}
 \langle \vec{\xi}^{\, 2}\rangle  =  
  \left(p_+v_+ \! + \! p_-  v_- \right)  N + \lambda^2.
\end{equation}

\subsubsection{Mathematical analysis}  \label{math} \ \\
We briefly recapitulate
the theory of on-line learning
as it has been applied to LVQ
in stationary environments
and refer to \cite{jmlr} for details. 

The {\it thermodynamic limit} $N\to\infty$ is
instrumental in the following.
As one of the simplifying consequences 
we can 
neglect the terms  $\lambda^2$ in  
Eq. (\ref{xinorm}). 
Moreover, the limit $N\to\infty$ facilitates 
the following key steps which, eventually, yield an exact mathematical description
of the training dynamics in terms of ODE:

\noindent
{\bf (I)} {\bf Order parameters:} The 
large number of adaptive prototype components can be  characterized in terms of only very few quantities.
  The definition of these order parameters follows directly from the
   mathematical structure of the model:
 \begin{equation} \label{orderdef}
  R_{S\sigma}^\mu = \vec{w}_S^\mu \, \cdot \vec{B}_\sigma
 \quad \mbox{and~}
  Q_{ST}^\mu = \vec{w}^\mu_S \cdot \vec{w}_T^\mu
 \mbox{~~~for all~} \sigma,S,T \in \{-1,+1\}.\end{equation}
The index $\mu$ represents the number
  of examples that have been presented to the system. 
 Obviously,  $Q_{+\!+}^\mu,Q_{-\!-}^\mu$ and 
 $Q_{+\!-}^\mu=Q_{-\!+}^\mu$ relate to the 
 norms and  
 overlaps of prototypes,
 the $R_{S\sigma}^\mu$ specify projections
 onto the cluster vectors
 $\{\vec{B}_+,\vec{B}_-\}$. \\
 {\bf (II)} {\bf Recursion relations:}
 For the above introduced order parameters, 
 recursion relations 
  can be  derived directly
 from the learning algorithm (\ref{generic}): 
 $$
  N^{-1}
  \left(R_{S\sigma}^{\mu} - R_{S\sigma}^{\mu-1}\right) 
  = \eta \, f_S  \left( \vec{B}_\sigma\cdot\vec{\xi}^\mu -
      R^{\mu-1}_{S\sigma}\right) \mbox{~~and~}
  N^{-1}\left({Q_{ST}^{\mu} - Q_{ST}^{\mu-1}} \right)     =\ldots 
  $$
  \begin{equation} 
  \label{recursions} 
  \ldots  \eta \left[f_S  \left(\vec{w}^{\mu-1}_T
  \!\cdot\!\vec{\xi}^\mu\!-\!
       Q^{\mu-1}_{ST}\right) 
    +  f_T \left( \vec{w}^{\mu-1}_S\!\cdot\!\vec{\xi}^\mu\!-\!
      Q^{\mu-1}_{ST}\right)\right]  
      + \eta^2  f_S  f_T (\vec{\xi}^\mu)^2/N.
   \end{equation}
  The modulation function is denoted as $f_\pm$ here, omitting its arguments.  Terms of order ${O}(1/N)$ have been
  discarded; note that  $(\vec{\xi}^\mu)^2
  = {O}(N)$ according to  (\ref{xinorm}). 

\noindent
{\bf (III)} {\bf  Averages over the data:} Applying the central limit theorem  (CLT)
 we can perform the average over the random sequence of  
 independent examples.
 The current  input $\vec{\xi}^\mu$ enters the r.h.s.\ 
 of Eq.\ (\ref{recursions}) only through its length and 
 \begin{equation} \label{hb}
  h_S^\mu \, = \vec{w}_S^{\mu-1} \cdot \vec{\xi}^\mu  \mbox{~~and~~}
  b_\sigma^\mu \, = \vec{B}_\sigma \cdot \vec{\xi}^\mu.
 \end{equation}
Since the scalar products correspond to sums of 
many independent random
quantities, the CLT applies and 
the projections in Eq.\ (\ref{hb})   
 are correlated Gaussian quantities 
 for large $N$.
Hence, 
their joint density 
is fully specified by 
the moments
 $$
 \left\langle h^\mu_{S} \right\rangle_{\sigma} = \lambda
 R_{S\sigma}^{\mu-1},
 \quad
  \left\langle b^\mu_{\tau} \right\rangle_{\sigma} = \lambda
  \delta_{S\tau}, \quad 
  \left\langle h^\mu_{S} h^\mu_{T} \right\rangle_{\sigma} -
  \left\langle h^\mu_{S} \right\rangle_{\sigma}
  \left\langle  h^\mu_{T} \right\rangle_{\sigma} = 
  v_\sigma \, Q^{\mu-1}_{ST}
 $$
  \begin{equation} \label{moments}
    \left\langle h^\mu_{S}  b^\mu_{\tau} \right\rangle_{\sigma} -
  \left\langle h^\mu_{S} \right\rangle_{\sigma}
  \left\langle  b^\mu_{\tau} \right\rangle_{\sigma} = 
  v_\sigma \, R^{\mu-1}_{S\tau}, \quad 
   \left\langle b^\mu_{\rho} b^\mu_{\tau} \right\rangle_{\sigma} -
  \left\langle b^\mu_{\rho} \right\rangle_{\sigma}
  \left\langle  b^\mu_{\tau} \right\rangle_{\sigma} = v_\sigma \, \delta_{\rho\tau}
  \end{equation}
 \ \\
 \noindent  where
   $\delta_{\ldots}$ is the Kronecker-Delta.  
  The joint density is therefore
  fully specified
  by the order parameters 
  of the previous time step and 
  by the model parameters $\lambda,p_\pm,v_\pm$. 
  This enables us to perform an average of the recursions (\ref{recursions}) over the latest example in terms
  of elementary Gaussian integrations. 
  Moreover, the result is obtained in
  in closed form in
   $\{ R_{S\sigma}^{\mu-1},Q_{ST}^{\mu-1} \} $,
   see \cite{jmlr} for details.
   
\noindent 
{\bf (IV)} {\bf Self-averaging properties} of the
 order parameters allow us to restrict the description 
 to their mean values, see \cite{reents} for a 
 mathematical discussion
  in the specific context of 
 on-line learning. Random flucutations vanish
 as $N\to\infty$ and, as a consequence, Eq.\   
 (\ref{recursions}) correspond to the
 deterministic
 dynamics of means. 
 
\noindent 
{\bf (V)} {\bf Continuous time limit and
learning curves:}
 For large $N$,
  we can interpret the ratios on the
  left hand sides of Eq.\ (\ref{recursions}) as derivatives
  with respect to the continuous learning time
  $ \alpha \, = {\, \mu\, }/{N}.$
  This corresponds to the natural expectation that  the number of examples required for successful training 
  should be proportional to the number of degrees of freedom 
  in the system.   
  The set of coupled ODE obtained
  from Eq.\ (\ref{recursions}) is of the generic form (see \cite{jmlr} for details)
  \begin{eqnarray}  \label{odegeneric} 
   \left[{dR_{S\tau}}\big/{d\alpha} \right]\,
    & = & 
    \eta \, \left( \left\langle b_\tau  f_S \right\rangle -  
    R_{S\tau}  \left\langle f_S \right\rangle \right) \\
  \left[{dQ_{ST}}\big/{d\alpha}\right]   & =&     \eta \, \left(
     \left\langle h_S f_T  + h_T  f_S \right\rangle
    -  Q_{ST}  \left\langle f_S  + f_T \right\rangle \right)
   + \eta^2 \,  {\textstyle \sum_{\sigma=\pm1} v_\sigma p_\sigma }
   \left\langle f_S  f_T \right\rangle_\sigma.
   \nonumber 
  \end{eqnarray} 
 The (numerical) integration yields the temporal
  evolution of order parameters in the 
  course of training. 
We consider prototypes
initialized as independent random vectors of squared
norm $\hat{Q}$ with no prior knowledge of the
cluster structure: 
\begin{equation}  
 Q_{++}(0)=Q_{--}(0)=\hat{Q}, \mbox{}
Q_{+-}(0)=0 \mbox{~and~} 
\label{initialconditions}
R_{S\sigma}(0)=0 \mbox{~for~} S,\sigma=\pm1.
\end{equation} 
The success of training is quantified 
in terms of the generalization error, i.e.\ the probability for
misclassifying novel, random data. 
Under the assumption that the density 
(\ref{prob})
represents the actual target
classification, we can work out 
the class-specific
errors for data from cluster 
$\sigma=1$ or $\sigma=-1$:
\begin{equation} \label{eg} \qquad
 \epsilon  \, =  \,  p_+ \, \epsilon^+  \, + \, p_- \, \epsilon^-  
  \mbox{~~~~with~~~}  \epsilon^\sigma \, = \,   \left\langle 
     \Theta\left( d_{+\sigma} - d_{-\sigma} \right) 
   \right\rangle_\sigma. 
\end{equation}
For the full derivation of the 
conditional averages 
as functions of order parameters
we  refer to \cite{jmlr}.
Exploiting self-averaging properties (IV)
again, 
we obtain the learning curve 
$\epsilon(\alpha)$, 
i.e.\ the performance after 
presenting $(\alpha \, N)$ examples. 

\vspace{-2mm}
\subsubsection{Weight decay} \ \\
Next, we extend the LVQ1 update by a so-called 
\textit{weight decay} term
 as an element of explicit 
\textit{forgetting.}  
To this end, we consider the
multiplication  of
all prototype components by a factor $(1-\gamma/N)
$ before the
generic learning step 
(\ref{generic}): 
\begin{equation}  \label{withdecay}
  \vec{w}_S^\mu \, 
    = \,  \left(1-{\gamma}\big/{N}\right) \,  \vec{w}_S^{\mu-1} \, +  \Delta \vec{w}_S^\mu.
 \end{equation}
 Since the multiplications with $\left(1-\gamma/N
 \right)$ 
 accumulate in the course of training, weight 
 decay results 
 in an increased influence of the most recent 
 training data
 as compared to {\it earlier} examples. 
 Similar modifications of 
 perceptron training in the presence
 of drift were  discussed in 
 \cite{drifting2,caticha2}.
 
  Other motivations for the introduction of 
  weight decay in machine learning range
 from the modelling of 
 {\it forgetful memories} in attractor neural networks \cite{mezard,hemmen}
 to {regularization} in order to reduce 
 over-fitting \cite{Hastie}.
 As an example for the latter,  weight decay in layered 
 neural networks 
 was analysed in \cite{saadsollawd}.

 The modified ODE for LVQ1 training with 
 weight decay, cf. Eq. (\ref{withdecay}), 
 are obtained in a straightforward manner and read
  \begin{equation}  \label{odedriftwd}  
   \left[{dR_{S\tau}}\big/{d\alpha} \right]_{\gamma} = \!
   \left[{dR_{S\tau}}\big/{d\alpha} \right] - \gamma R_{S\tau}
  \mbox{;~~} 
   \left[{dQ_{ST}}\big/{d\alpha}\right]_\gamma =\!
  \left[{dQ_{ST}}\big/{d\alpha}\right] - 2\, \gamma \,Q_{ST} 
  \end{equation}
 with the terms $[\ldots]$ on the r.h.s.\ 
 formally given by Eq.\ (\ref{odegeneric})
 for $\gamma=0$.   

\subsection{LVQ dynamics under concept drift}  \label{underdrift} 

The analysis summarized in the previous section 
concerns 
learning in stationary environments with  densities and targets of
the form (\ref{prob}). Here, we discuss the effect of including concept
 drift  within our modelling 
framework. \vspace{-5mm}
\subsubsection{Real drift of the
 cluster centers} \ \\
In the presented framework, a real drift can
be modelled
by processes that displace the 
cluster centers in $N$-dim.\ feature space
while the training follows the stream of
data.  
As a specific example, in
\cite{preprint}  the authors study
 the effects of a 
 random {\it diffusion} of time-dependent
 vectors $\vec{B}_\pm (\mu)$.
 The results show that simple
 LVQ1 is capable of tracking randomly drifting concept to a non-trivial
 extent.  
 \vspace{-5mm} 
 
\subsubsection{Time dependent input densities}\ \\
Strictly speaking, virtual drifts 
affect only the statistical properties
of observed example data, while 
the actual target classification remains
the same. 
In our modelling framework, we 
can readily consider 
time-de\-pen\-dent parameters of the density
(\ref{prob}), e.g. $\lambda(\alpha)$ and
$v_\sigma(\alpha)$ by inserting them 
in Eq. (\ref{odegeneric}). 
\begin{figure}[t]
\mbox{~~}
\includegraphics[width=0.5\textwidth]{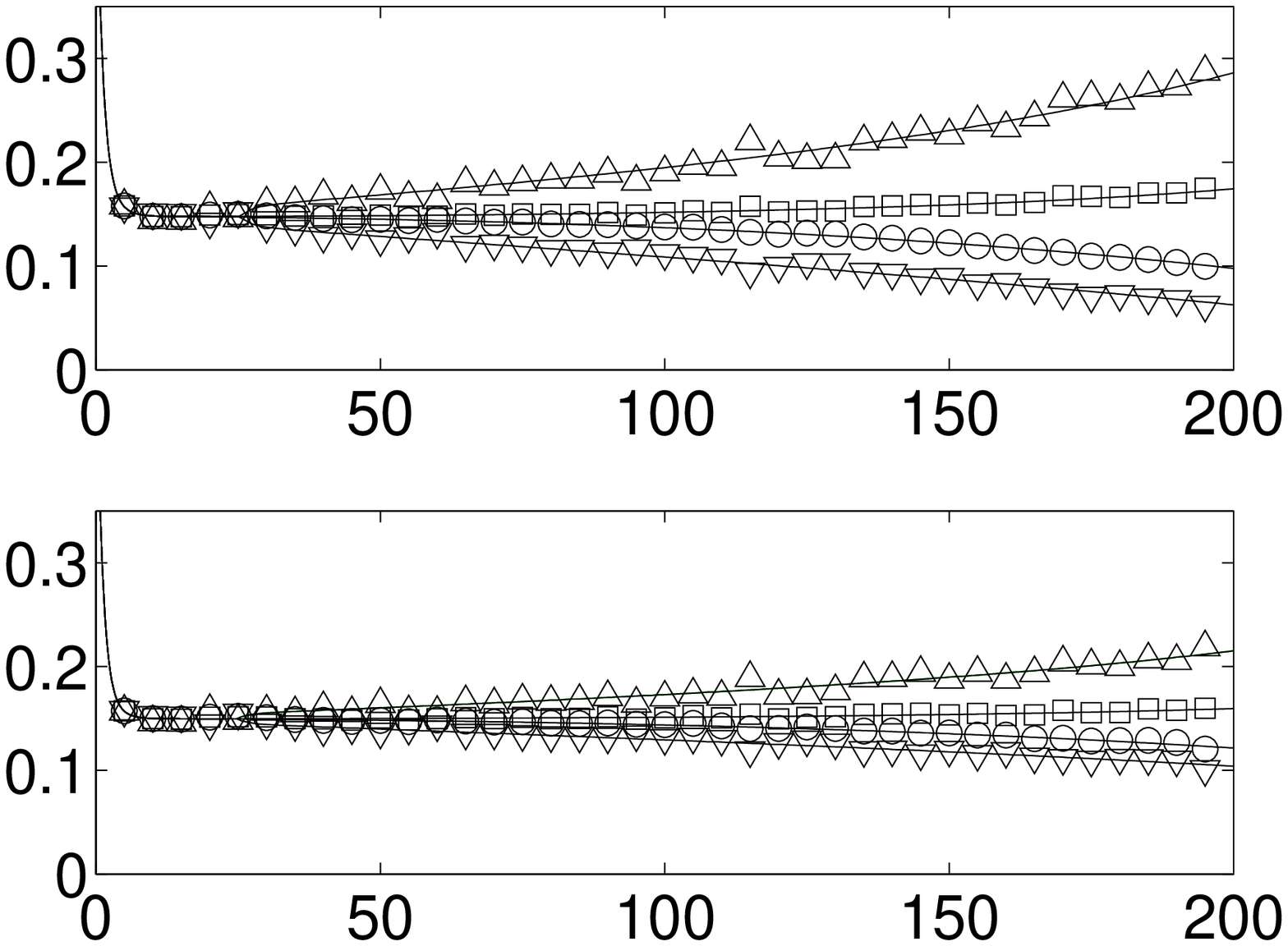}
\put(-45,8){\Large $\alpha$}
\put(-45,-52){\Large $\alpha$}
\put(-175,52){\Large $\epsilon$}
\put(-175,-8){\Large $\epsilon$}
\includegraphics[width=0.5\textwidth]{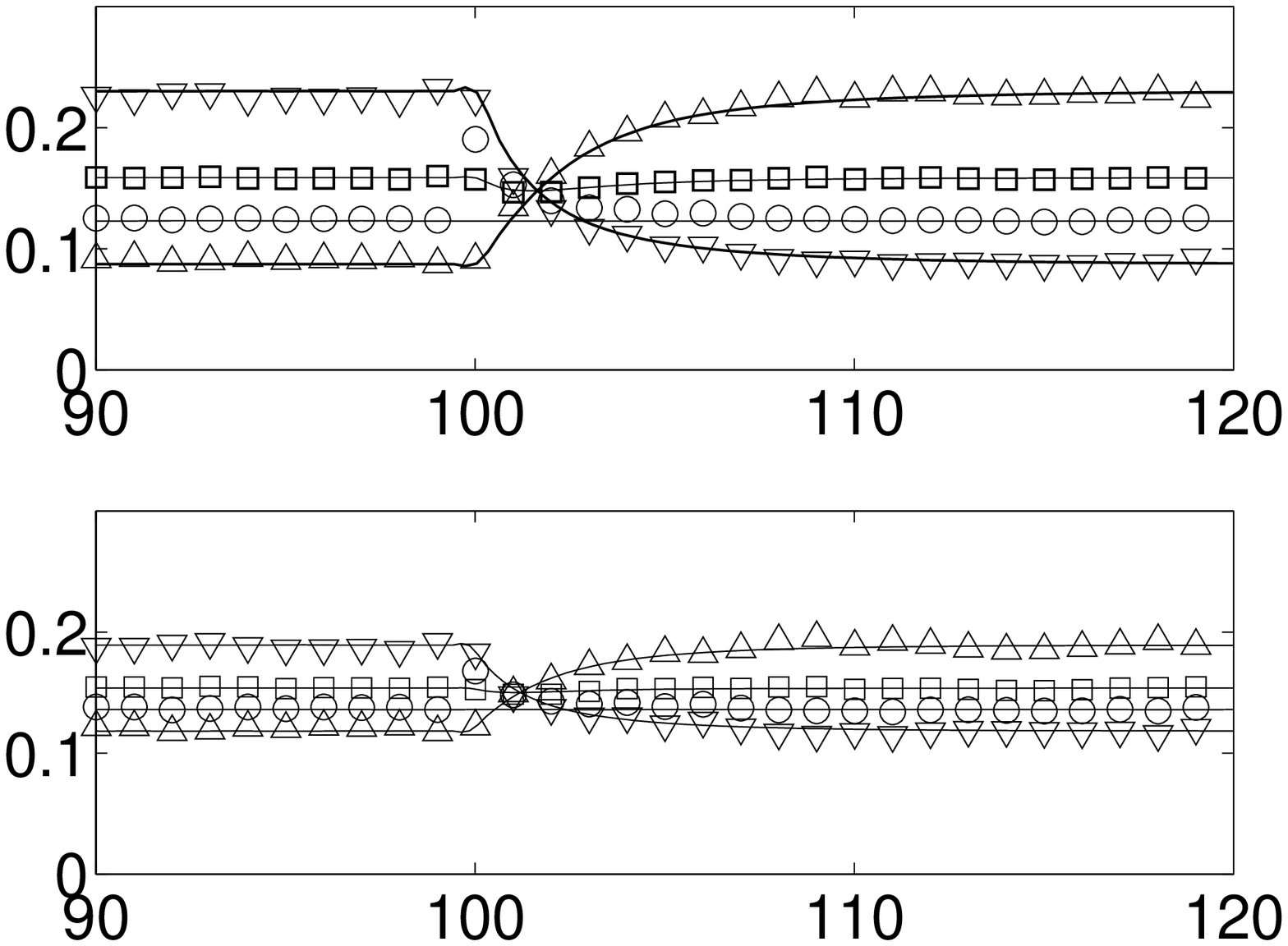}
\put(-45,8){\Large $\alpha$}
\put(-45,-52){\Large $\alpha$}
\put(-175,52){\Large $\epsilon$}
\put(-175,-8){\Large $\epsilon$}
\caption{ \label{linplot} 
LVQ1 in the presence of concept drift. 
Solid lines correspond to the integration of
ODE with
initialization as in Eq.\ 
(\ref{initialconditions}). 
Cluster variances are $v_+\!=\!v_-\!=\!0.4$
and $\lambda=1$ in the density (\ref{prob}). 
Upper graphs correspond to LVQ1 without 
weight decay, lower graphs display results for
 $\gamma=0.05$ 
in (\ref{withdecay}). In addition, 
Monte Carlo results for $N=100$ are shown: 
class-wise errors $\epsilon^\pm(\alpha)$  
are displayed as upward (downward) triangles, 
respectively;  
 squares mark 
 the reference error $\epsilon_{ref}(\alpha)$ 
 (\ref{epsref}); circles correspond to 
 $\epsilon_{track}(\alpha)$ 
 (\ref{epstrack}).
{\bf Left panel:} drift with 
linearly increasing $p_+(\alpha)$ given by 
$\alpha_o\!=\!20$,
$\alpha_{end}\!=\!200$, $p_{max}\!=\!0.8$
in (\ref{linincrease}). 
 {\bf Right panel:} sudden change of class 
weights according to
to Eq.\ (\ref{sudden}) with 
$\alpha_o=100$  and $p_{max}=0.75$. 
Only the $\alpha$-range close
to $\alpha_o$ is shown.
}
\end{figure}
Here, we will focus on  non-stationary 
prior weights $p_+(\alpha)=1-p_-(\alpha)$ 
for the generation of example data.  
In this case,
a varying fraction of examples
represents each of the classes in the stream
of training data. 
Non-stationary class bias
complicates the training
significantly and  can lead to inferior performance
in practical situations
\cite{Wang}.

 \paragraph{(A) Drift in the training data only} \ \\
 Here we assume 
 that the target classification
 is defined by a fixed 
 \textit{reference density}
 of data. 
 As a  simple example we consider
 equal priors $p_+=p_-=1/2$ in a  
 symmetric reference density
 (\ref{prob}) with $v_+=v_-$.  
 On the contrary, the  characterstics 
 of the observed 
 training data is assumed to be
 time-dependent. In particular, we study 
 the effect of time-dependent
 $p_\sigma(\alpha)$ and weight decay. 
 
 Given the order parameters of the learning 
 systems in the course of training,
 the corresponding \textit{reference
 generalization error}
 \begin{equation} \label{epsref}
 \epsilon_{ref}(\alpha)=
 \left(\epsilon^+ +  \epsilon^-\right)/2
 \end{equation} 
 is obtained
 by setting $p_+=p_-=1/2$
 in Eq. (\ref{eg}), but inserting 
 $R_{S\tau}(\alpha)$ and $Q_{ST}(\alpha)$
 as obtained from the integration of the ODE
 (\ref{odegeneric}) or  (\ref{odedriftwd}) 
 with time dependent
 $p_+(\alpha)=1-p_-(\alpha)$ in the training 
 data.

 \paragraph{(B) Drift in training and test data} \ \\
 In the second interpretation we assume that the 
 time-dependence of $p_\sigma(\alpha)$ 
 affects both the 
 training and  test data in the same way.  
 Hence, the change of the 
 statistical properties of
 the data is inevitably accompanied 
 by a modification of  the target 
 classification: For instance, 
 the  Bayes optimal classifier and
 its best linear approximation
 will depend explicitly on the current priors
  $p_\sigma(\alpha)$ \cite{jmlr}.

 The learning system is supposed to 
 track the drifting concept and we denote
 the corresponding generalization error, cf. 
 Eq. (\ref{eg}),
 by 
 \begin{equation} \label{epstrack} \epsilon_{track}=
 p_+(\alpha) \epsilon^+ 
 + p_-(\alpha) \epsilon^-.
 \end{equation}  
 
 In terms of modelling the training dynamics, 
 both scenarios, (A) and (B), require the 
 same straightforward
 modification of the ODE system: the
 explicit introduction of $\alpha$-dependent
  quantities $p_\sigma(\alpha)$ 
 in Eq.\ (\ref{odegeneric}). However, the obtained
 temporal evolution translates into the 
 reference error $\epsilon_{ref}(\alpha)$ for
 the case of drift in the training data (A), and into
 $\epsilon_{track}(\alpha)$ in 
 interpretation (B). 

\section{Results and Discussion} \label{results}

Here we present and discuss 
first results obtained by 
integrating the systems of ODE for LVQ1 
with and without weight decay under 
different time-dependent drifts. For comparison,
averaged learning curves as obtained by means
of Monte Carlo simulations are also shown.
All results are for constant
learning rate $\eta=1$ and the LVQ systems
were initilized according to Eq.\ 
(\ref{initialconditions}).

We study three example scenarios for the time-dependence $p_+(\alpha)=1\!-p_-(\alpha)$: 

\noindent 
{\bf Linear increase of the bias} \\
We consider a time-dependent bias of the form
$ p_+(\alpha) = 	 
     1/2  \mbox{~for~~} \alpha<\alpha_o$ and  
     \begin{equation} \label{linincrease} 
     p_+(\alpha) =
     1/2 + \frac{(p_{max}\!-\!1/2) \, (\alpha-\alpha_o)}{(\alpha_{end}-\alpha_o)}  
     \mbox{~for~~} \alpha\geq\alpha_o.
 \end{equation} 
 where the maximum class weight $p_+=p_{max}$ 
 is reached at learning time $\alpha_{end}$. 
Fig.\ \ref{linplot} (left panel) shows the 
learning curves as obtained by numerical
integration of the ODE together with Monte Carlo
simulation results for $(N=100)$-dimensional
inputs and prototype vectors. 
As an example we set the parameters to
$ \alpha_o=25, p_{max}=0.8, \alpha_{end}=200$.
The learning curves are displayed for 
LVQ1 without weight decay (upper) and with $\gamma=0.05$ (lower panel).
Simulations show excellent
agreement with the ODE results.

The system adapts to the increasing imbalance
of the training data, as reflected by a
decrease (increase) of the class-wise error for 
the
over-represented  (under-represented) class,
respectively. 
The weighted over-all error
 $\epsilon_{track}$ also decreases, i.e. the 
 presence of class bias facilitates smaller
 total generalization error, see \cite{jmlr}. 
 The performance with respect to unbiased
 reference data detoriorates slightly, i.e.
 $\epsilon_g$ grows with increasing class bias
 as the training data represents the target
 less faithfully. 
 
 The influence of the class bias and its 
 time-dependence is reduced significantly in
 the presence of weight decay with 
 $\gamma>0$,
 cf. Fig. \ref{linplot} (lower panel). 
 Weight decay restricts the norm of the 
 prototypes, i.e. the possible offset 
 of the decision boundary from the origin. 
 Consequently, the tracking error slightly
 increases,  while $\epsilon_{ref}$  with
 respect to the reference density is
 decreased compared to the setting without
 weight decay, respectively.

\noindent 
{\bf Sudden change of the class bias}\ \\
Here we consider an instantaneous switch
from   high bias $p_{max} >1/2$ to
 low bias: 
\begin{equation} \label{sudden} 
  p_+(\alpha) = 	 
     p_{max} \mbox{~for~~} \alpha< \alpha_o 
     \mbox{~~and~~} p_+(\alpha)= 
     1 -p_{max} \mbox{~for~~} \alpha\geq \alpha_o.
 \end{equation} 
We consider $p_{max}=0.75$ as an example, 
the corresponding results from the integration
of ODE and Monte Carlo simulations are shown
in Fig.\ \ref{linplot} (right panel) for
training without weight decay (upper)
and for $\gamma=0$ (lower panel). 

 \begin{figure}[t!]
 \mbox{~~}
\includegraphics[width=0.48\textwidth]{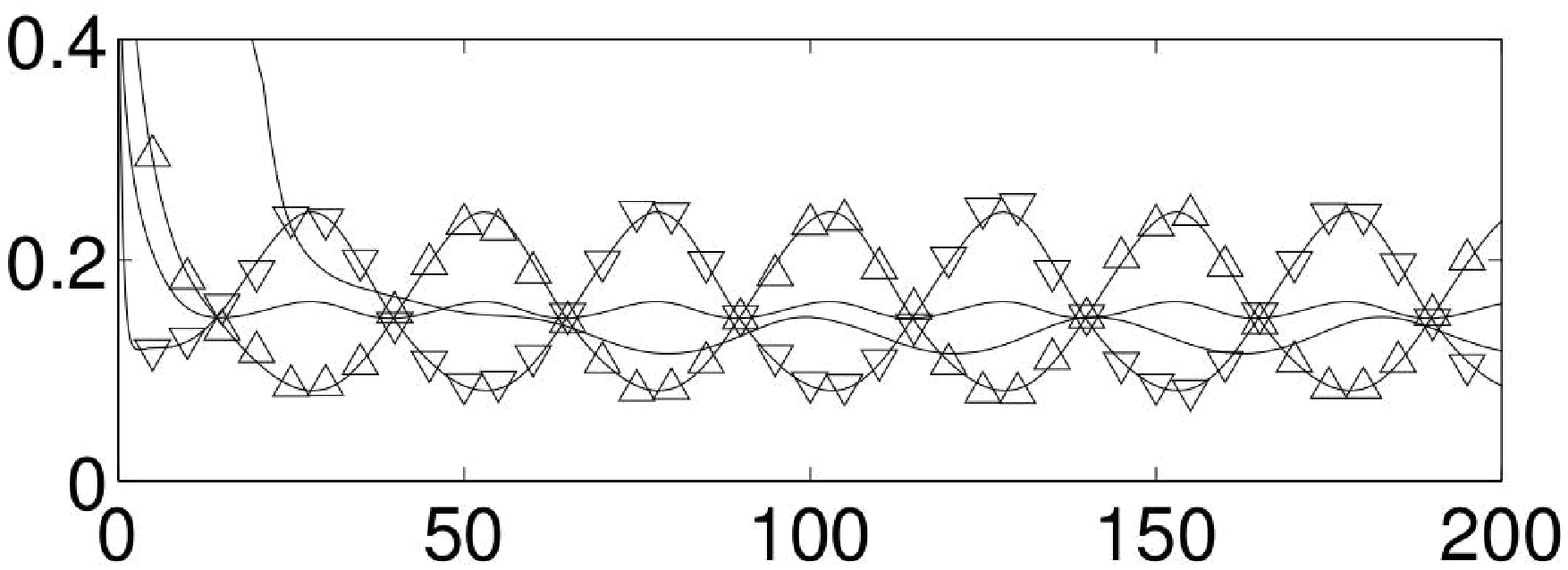}
\put(-30,0){\Large $\alpha$}
\put(-170,40){\Large  $\epsilon$}
\mbox{~~}
\includegraphics[width=0.48\textwidth]{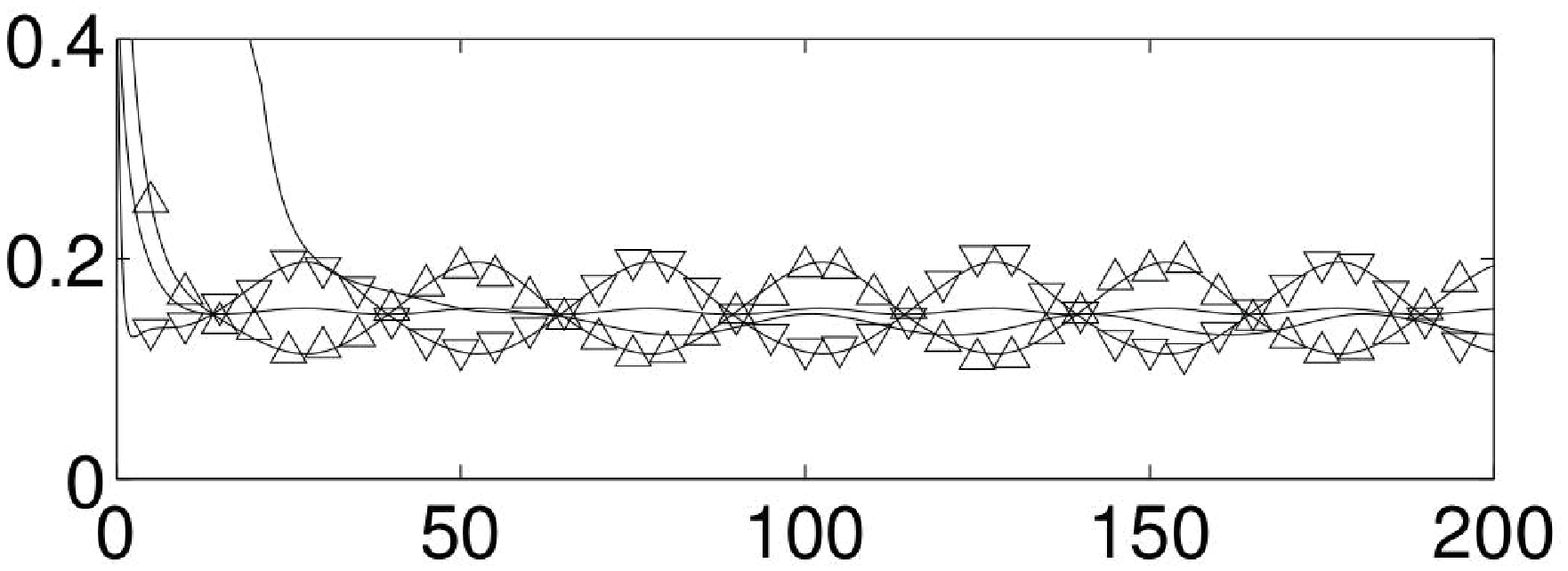}
\put(-30,0){\Large $\alpha$}
\put(-170,40){\Large  $\epsilon$}
\caption{ \label{periodicplot} 
LVQ1 in the presence
of oscillating class 
weights according
to Eq.\ (\ref{oscillating}) with parameters
$T=50$  and $p_{max}=0.8$, 
without weight decay $\gamma=0$ (left)
and for $\gamma=0.05$ (right). 
For clarity, Monte Carlo results are only shown
for the class-conditional errors $\epsilon^+$ 
(upward) and $\epsilon^-$ (downward triangles). 
All other settings as in  Fig.\  \ref{linplot}.} 
\end{figure}

We observe similar effects as for the slow,
linear time-dependence: The system reacts
rapidly with respect to the class-wise errors 
and the tracking error $\epsilon_{track}$ 
maintains a relatively low value. Also, the
reference error $\epsilon_{ref}$ displays 
robustness with respect to the sudden change
of $p_+$.  Weight decay, 
as can be seen in the lower right panel of Fig.\ref{linplot} reduces the over-all
sensitivity to the bias and its change: 
Class-wise errors are more balanced and the
weighted $\epsilon_{track}$ slightly increases
compared to the setting with $\gamma=0$.

The weight decay does not seem to have a
 notable effect on the promptness of
 the system's adaptation to the changing
 bias. While it significantly regularizes
 the system, the expected effect of 
 \textit{forgetting} previous information
 in favor of the most recent examples cannot
 be observed.

 \noindent 
{\bf Periodic time dependence} \ \\
As a third scenario we consider an oscillatory
modulation of the class weights in training:  
\begin{equation} \label{oscillating}  
  p_+(\alpha) = 	 1/2 + (p_{max}-1/2) \, \cos\left[ 2\pi \, 
  {\alpha}\big/{T} \right]
 \end{equation} 
 with periodicity $T$ on $\alpha$-scale and
 maximum amplitude $p_{max} <1$.
  
 Example results are shown in Fig.\ \ref{periodicplot} for $T=50$ and 
 $p_{max}=0.8$. Monte Carlo results for $N=100$
 are only displayed for the class-wise errors
 show excellent agreement with the 
 numerical integration of the ODE for training
 without weight decay (left panel) and for 
 $\gamma=0.05$ (right panel).
 The observations confirm our findings for
 slow and sudden changes of $p_+$: In the main,
 weight decay limits the reaction of the
 system to the presence of a bias and its
 time-dependence. 
 
\vspace{-2mm}
\section{Summary and Outlook} 
\vspace{-2mm}
In summary, we have presented
a mathematical framework
in which to study the influence of concept drift
on prototype-based classifiers systematically. 
In all specific drift scenarios considered here,
we observe that
simple LVQ1 can track the time-varying class
bias to a non-trivial extent:
In the
interpretation of the results in terms
of real drift, the class-conditional performance
and the 
tracking error $\epsilon_{track}(\alpha)$ clearly 
reflect the time-dependence of the prior weights.

In general, the reference error 
$\epsilon_{ref}(\alpha)$ 
with respect to  class-balanced test data,  
displays only little
deterioration due to the drift in the
training data.

The main effect of introducing weight decay 
is  a reduced overall sensitivity to bias in the
training data: Figs.\ 1-3 display
a decreased difference between the class-wise
errors $\epsilon^+$ and $\epsilon^-$ for 
$\gamma>0$. 
Na\"ively, one might have expected
an improved tracking of the drift due to the
imposed \textit{forgetting}, resulting in, for instance, 
a more rapid reaction to the 
sudden change of bias in Eq.\  (\ref{sudden}).  
However, such
an improvement cannot be confirmed. 
Our findings are in contrast to a recent study
\cite{preprint}, in which we observe    
increased performance by weight
decay for a different drift scenario, i.e.\ 
the randomized displacement of
cluster centers. 

The precise
influence of weight decay clearly 
depends on the
geometry and relative position of the clusters.
Its dominant effect, however, is the 
regularization
of the LVQ system by 
reducing the norms $Q_{++}$ and $Q_{--}$ 
of the prototype vectors.
Consequently, the NPC classifier
is less flexible to reflect 
class bias which would require
significant offset
of the prototypes and decision boundary 
from the origin. This mildens the influence
of the bias (and its time-dependence) and
results in a more robust behavior of the
employed error measures.

 Alternative
mechanisms of \textit{forgetting} should
be considered which do not limit the flexibility
of the LVQ classifier, yet facilitate 
\textit{forgetting} of older information. 
As one example strategy  we intend to
 investigate 
the accumulation of additive 
noise in the training process. 
We will also explore the parameter space of
the model density and in greater depth and 
study the
influence of the learning rate systematically. 
\vspace{-3mm}
 \subsubsection{Acknowledgement} \ \\
The authors would like to thank A. Ghosh, 
A. Witoelar and G.-J. de Vries
for useful discussions of earlier projects on
LVQ training in stationary environments.

\bibliographystyle{unsrt}  
\bibliography{biehl-drift}

\end{document}